\documentclass[lettersize, 10 pt, journal, twoside]{IEEEtran}
\usepackage{amsmath,amsfonts}
\usepackage{algorithmic}
\usepackage{array}
\usepackage[full]{textcomp}
\usepackage{stfloats}
\usepackage{url}
\usepackage{verbatim}
\usepackage{graphicx}
\usepackage{subcaption}
\def\BibTeX{{\rm B\kern-.05em{\sc i\kern-.025em b}\kern-.08em
    T\kern-.1667em\lower.7ex\hbox{E}\kern-.125emX}}
\usepackage{balance}

\IEEEpubid{\copyright~2022 IEEE}
\usepackage{cite}
\usepackage{amsmath,amssymb,amsfonts}
\usepackage{algorithmic}
\usepackage{graphicx}
\usepackage{dsfont}
\usepackage{xcolor}

\usepackage[font=small,skip=3pt,tableposition=top]{caption}
\usepackage{mathptmx} % assumes new font selection scheme installed
\usepackage{times} % assumes new font selection scheme installed

\usepackage{url}
\usepackage{diagbox}
 
\usepackage{amsfonts, amssymb}

\usepackage{newtxtext, newtxmath}
\usepackage[export]{adjustbox}
\usepackage{bbm}

\usepackage{mathtools}
 
\newcommand{\norm}[1]{\left\lVert#1\right\rVert}
\newcommand{\Var}[1]{\text{Var}}
\newcommand{\EXP}[1]{\mathop{\mathbb{E}}{}}

\usepackage[ruled,norelsize]{algorithm2e}
\usepackage{graphicx}
\usepackage{tabularx}
\usepackage{caption}
\usepackage{multirow}
\usepackage{standalone}
\usepackage{amsmath}
\usepackage{color}

\begin{document}
\title{ Hierarchical Policies for Cluttered-Scene Grasping with Latent Plans}

\author{Lirui Wang, Xiangyun Meng, Yu Xiang, Dieter Fox
\thanks{Manuscript received: September, 9, 2021; December, 4, 2021; Accepted January, 10, 2022.}%Use only for final RAL version
\thanks{This paper was recommended for publication by Editor Markus Vincze upon evaluation of the Associate Editor and Reviewers' comments.}
\thanks{Lirui Wang is with the Computer Science and Artificial
Intelligence Laboratory, Massachusetts Institute of Technology, USA. {\tt\small liruiw@mit.edu }} \thanks{Xiangyun Meng and Dieter Fox are with the Paul G. Allen School of Computer Science \& Engineering at the University of Washington, USA. {\tt\small\{xiangyun, fox\}@cs.washington.edu}. Dieter Fox is with NVIDIA, USA. {\tt\small dieterf@nvidia.com }}
\thanks{Yu Xiang is with the Department of Computer Science at the University of Texas at Dallas, USA. {\tt\small yu.xiang@utdallas.edu}}
\thanks{Digital Object Identifier (DOI): see top of this page.}
} 

\markboth{IEEE Robotics and Automation Letters. Preprint Version. January, 2022}
{Wang \MakeLowercase{\textit{et al.}}: Hierarchical} 
% make the title area

\maketitle

%%%%%%%%%%%%%%%%%%%%%%%%%%%%%%%%%%%%%%%%%%%%%%%%%%%%%%%%%%%%%%%%%%%%%%%%%%%%%%%%

\begin{abstract}
6D grasping in cluttered scenes is a longstanding problem in robotic manipulation. Open-loop manipulation pipelines may fail due to inaccurate state estimation, while most end-to-end grasping methods have not yet scaled to complex scenes with obstacles. In this work, we propose a new method for end-to-end learning of 6D grasping in cluttered scenes. Our hierarchical framework learns collision-free target-driven grasping based on partial point cloud observations. We learn an embedding space to encode expert grasping plans during training and a variational autoencoder to sample diverse grasping trajectories at test time. Furthermore, we train a  critic network for plan selection and an option classifier for switching to an instance grasping policy through hierarchical reinforcement learning.  We evaluate our method and compare against several baselines in simulation, as well as demonstrate that our latent planning can generalize to real-world cluttered-scene grasping tasks.\footnote{Video and code can be found at \url{https://sites.google.com/view/latent-grasping}.}
\end{abstract}
\begin{IEEEkeywords}
Deep Learning in Grasping and Manipulation; Sensorimotor Learning
\end{IEEEkeywords}
\IEEEpeerreviewmaketitle

\section{Introduction}

\label{sec:intro}
\IEEEPARstart{6}D robotic grasping of a specific object in cluttered scenes is important for common robotic applications such as a  robot trying to pull out a cereal box from a shelf without moving other objects. Recently, there has been significant progress on using deep learning for grasp synthesis \cite{redmon2015real,mahler2017dex,mousavian20196} and for policy-based methods \cite{kalashnikov2018qt,wang2021goal,song2019grasping} that directly output end-effector actions. However, the former approach heavily relies on the perception system and treats motion generation as a separate problem, while the latter approach has not scaled to challenging settings with obstacles. For long-horizon and multi-modal grasping tasks \cite{nakamura2017complexities}, we propose that end-to-end mapping from perception to low-level control can benefit from a latent representation of grasping trajectories.

% We consider target-driven grasping in clutter scenes \cite{murali20196} where collisions and unintended contact are more catastrophic than picking an object indiscriminately in a pile. In a standard motion and grasping planning pipeline \cite{kitaev2015physics,dogar2011framework,wang2019manipulation}, the model-based planners often operate  good state estimation, and only consider information and metrics before execution. This modularity ignores the connection among perception and planning and control and thus suffers from the open-loop errors \cite{kappler2018real}. This is especially true for cluttered scenes due to the heavy occlusions and potentials for collisions.

We consider the target-driven grasping problem in clutter scenes \cite{murali20196} where collisions and unintended contact must be avoided. In a standard motion and grasp planning pipeline \cite{kitaev2015physics,dogar2011framework,wang2019manipulation}, model-based planners are often sensitive to state estimation errors in object poses and shapes, which may cause unwanted collisions. Recent deep grasp synthesis methods \cite{mahler2017dex,mousavian20196} address some of these limitations, but still execute motion plans in an open loop that suffers from occlusion due to partial observability \cite{kappler2018real}. This open-loop separation does not exploit the connection among perception, planning and control. %, which results in sometimes makes grasping difficult for cluttered scenes due to the heavy occlusions and potentials for collisions.
Furthermore, most learning-based methods on closed-loop grasping \cite{kalashnikov2018qt,wang2021goal,song2019grasping} are restricted to output  immediate low-level actions such as joints or end-effector poses, thus they fail to account for the \emph{multi-modality} nature and the \emph{temporal} structure in cluttered-scene grasping.
%often lack the structures needed for complex scenarios that involve \emph{multi-modality}, and are limited to handle bin picking or instance grasping without obstacles.

%  Under heavy occlusions, pose estimation and scene understanding are difficult. A grasp sampler might be able to provide diverse grasp sets and acquires plans, but the metric for selecting a specific grasp and plan often only depends on metrics independent of the execution. 

\begin{figure*}
\centering 
\includegraphics[width=0.65\linewidth ]{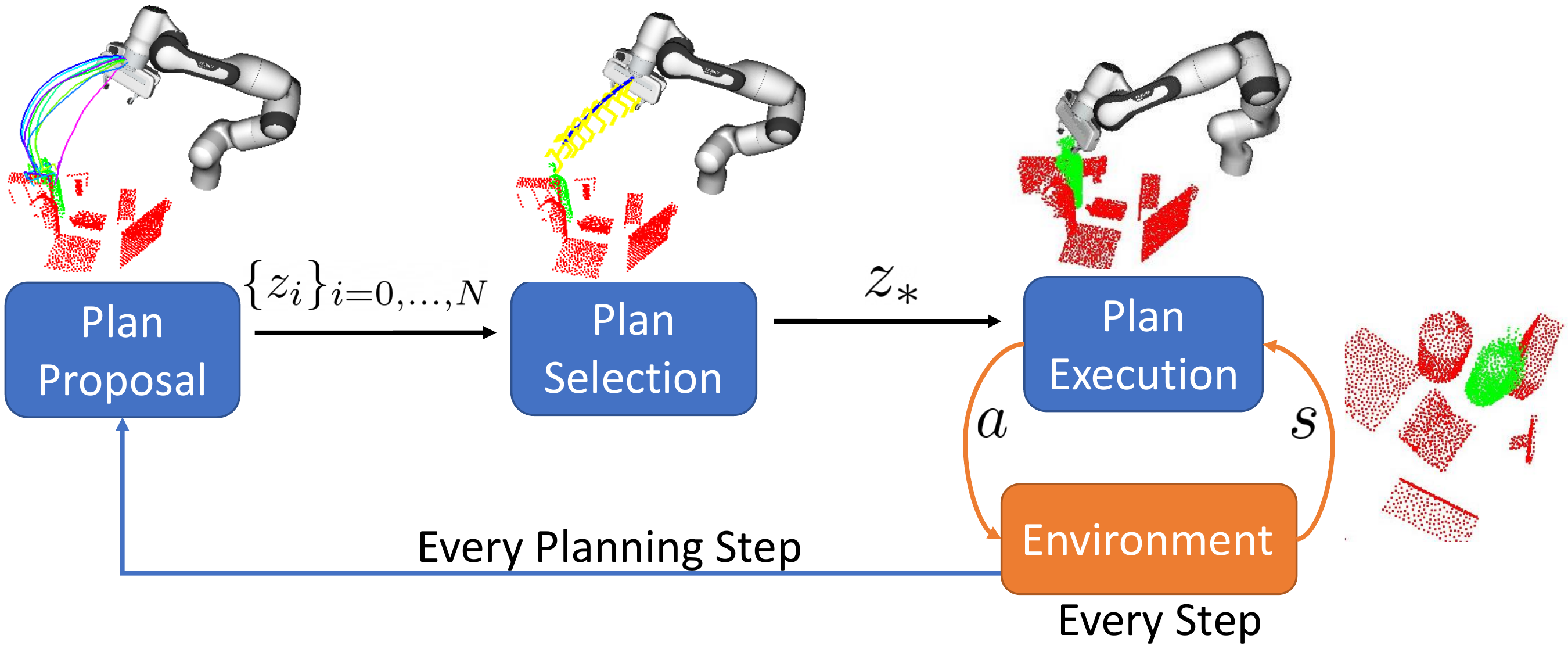}
  \caption{Illustration of the HCG framework.   
 Our method learns a closed-loop hierarchical policy for cluttered scene grasping from segmented point clouds (green denotes target and red denotes obstacles). We learn a high-level latent plan sampler and a latent plan critic network to generate and select optimal plan. We then use a low-level plan-conditioned policy and a grasping primitive (at a higher frequency) to avoid obstacles and execute grasping. The latent plans (colored lines) are visualized by recursively applying policy actions to the point clouds at current timestep in closed loop. }
  \label{fig:mpc}
  \vspace{-2mm}
\end{figure*}

% Illustration of the HCG framework.   
%  Our method learns a hierarchical policy to handle the multi-modality of cluttered scene grasping from segmented point clouds (green denotes target and red denotes obstacles). We represent high-level nominal plans as latents and learn a latent plan sampler to generate diverse plans. We then use a critic network to select among plans and use an option for single-object grasping primitive. Finally our plan-conditioned low-level policy executes the selected plan in closed-loop. The plans (colored lines) are visualized by recursively applying low-level actions to the observed point clouds, which can be treated as model-based rollouts in a deterministic and noise-free robot dynamics model.
 
To bridge the gap between traditional open-loop planning pipelines and learning-based reactive grasping methods, we propose to use a hierarchical policy to plan in the latent space \cite{hausman2018learning,lynch2020learning}. The planning is done via sampling and scoring latent trajectory embeddings. To capture the multi-modality in cluttered-scene grasping, we learn a generative model (the high-level policy) from a large number of expert demonstrations of grasping in cluttered scenes. At test time, given a partially-observed point cloud of a new scene, the high-level policy outputs latent trajectory embeddings that encode diverse plausible trajectories. The embeddings are then scored via a critic network which is trained via Q-learning. The low-level policy takes the highest-scored embedding and outputs a sequence of end-effector poses for the robot to follow. Finally, to improve the final grasping success rate, we switch from trajectory following to a reactive policy \cite{wang2021goal} at the end. The switching happens autonomously via learning a switching classifier jointly with the critic network.

%can exploit the multi-modality through considering the diverse plans rather than sticking to reach a single fixed configuration. Therefore, we represent the multi-modal grasping plans as latent codes and use latent plan sampling and selection mechanism to improve over the local nature of fixed-endpoint planning and low-level control.

% Specifically, we adopt a workspace low-level control policy that takes partially-observed point clouds of objects as input. Given a latent vector that summarizes a nominal plan, the low-level policy is trained to avoid obstacles and approach the target. To compute a feasible latent plan, we train a generative model to sample latent plans through variational inference and a Q network to select latent plan through hierarchical reinforcement learning (HRL). Additionally, we jointly train a policy switch classifier to use a single-instance 6D grasping policy \cite{wang2021goal} for the final grasping stage. Our method reactively plans and controls for cluttered-scene grasping at test time, i.e. sample plans with the high-level policy, choose a plan with the high-level critic, execute the selected plan with the low-level policy, and close the loop, shown in Fig.~\ref{fig:mpc}.  

We conduct thorough analyses and evaluations on the components and design choices of our hierarchical method. We show that the latent embeddings learned from expert plans capture the multi-modal nature of grasping trajectories in cluttered scenes. At test-time, the latent sampler and critic can generate and select plans to avoid obstacles and execute accurate grasping. Our method also outperforms several traditional manipulation baselines both in efficiency and success rates in unseen simulation test scenes. Finally, We show our method can successfully avoid obstacles and achieve high grasp success rates in cluttered real-world scenarios.

% We also conduct qualitative analysis of the latent space representation and ablate on the method components.
\IEEEpubidadjcol
Overall, our contributions are:
1) We introduce an end-to-end cluttered-scene 6D grasping learning framework based on target-centric masked point clouds. 2) We propose Hierarchical policies for Cluttered-scene Grasping (HCG) to factorize a grasping plan as a latent vector that can be sampled from a generative model and can be executed with a plan-conditioned  policy. We further improve the performance by learning a critic to select plans and a policy switch option \cite{wang2021goal} for the final grasping through online interactions. 3) We perform detailed experiments and validations in simulation and also show that our policy can be successfully applied to challenging real-world cluttered scenes.
\section{Related Work}
\label{sec:related}
\textbf{Learning-based Cluttered-Scene Grasping:}
Target-driven grasping in structured clutter scenes has been studied in traditional motion and grasp planning literature with no learning components~\cite{kitaev2015physics,dogar2011framework,wang2019manipulation}.
The vision-based grasping literature largely concerns either 6D grasping synthesis for a single isolated object followed by postprocessing and planning~\cite{pinto2016supersizing,mahler2017dex,morrison2018closing,murali20196} or planar grasping policy for bin-picking scenarios  \cite{levine2018learning,kalashnikov2018qt}. The former method includes state estimation with analytic grasp synthesis \cite{miller2004graspit} as well as direct grasp synthesis from observations \cite{bohg2013data}. For the latter approach, most works learn top-down grasping policies to grasp rather small and light objects that are spread in a bin~\cite{song2019grasping,mahler2017dex}. Instead, our method learns an object-centric 6D grasping policy in structured clutters, which can handle obstacle avoidance and multi-modal grasps. 6D grasping is also more flexible for partial views, diverse object poses, and additional tasks or arm constraints. Inspired by the recent works on state representations~\cite{wang2021goal,murali20196}, our approach uses accumulated segmented scene point clouds to represent target and obstacles and reduces the sim-to-real perception gap.

% A segmentation mask is used to specify target object and the policy needs to grasp and avoid obstacles.  Our work is related to grasp synthesis and traditional planning methods since our learning method is used to generate grasping trajectory.

% The challenging long-horizon grasping problem is then casted to a latent plan generation problem for the high-level policy, which is then solved with standard RL techniques.  
% To tackle this, our method learns a closed-loop MPC policy for spatial grasping in cluttered scenes.

\textbf{Off-Policy Learning and Hierarchical Policy:}
We consider a model-free approach to tackle the sparse-reward grasping problem.  Our work is related to policy methods that use demonstrations as priors to speed up learning and exploration~\cite{vecerik2017leveraging,nair2018overcoming}, as well as hierarchical policies~\cite{sutton1999between,bacon2017option} that learn temporally extended actions or options. Decomposing long-horizon tasks into skill abstractions or high-level actions is often shown to improve generalization and task transfer. Previous works~\cite{nachum2018data,nair2018visual,mandlekar2020iris} have used  subgoals or discrete actions as high-level actions.  Inspired by these methods, our work uses a motion and grasp expert planner~\cite{wang2019manipulation} to provide demonstrations. Furthermore, our method uses a high-level policy to generate abstract latent plans rather than goals for the low-level policy. We also use a binary classifier to choose a pretrained grasping policy option~\cite{wang2021goal} to improve the grasping success.

% The method can be treated as a model-free analogy of MPC \cite{agachi20162} systems that incorporate lookahead planning by reactively sampling and selecting plans. Without the critical goal requirements, we use latent variable to represent plans and uses standard Q learning to evaluate plans.

\textbf{Behavior learning with Latent Variables:}
 Latent variables have been used to represent complex skills in multi-task settings and to interpret the learned behavior semantics~\cite{hausman2018learning,lynch2020learning}. Since high-level behaviors are typically multi-modal, prior works have used generative models to model the latent behavior distribution. For instance, \cite{singh2020parrot} trains a transformed action space with RealNVPs \cite{dinh2016density}  and \cite{lynch2020learning} uses cVAE to learn latent plans from plays for a plan-conditioned policy. 
These works usually require goals as input at test time or only work with environments with limited scene variations and complexity such as contacts and obstacles. Instead, our method focuses on the context of cluttered-scene grasping and exploits the multi-modality of potential grasping plans. We use latent variables in a conditional VAE \cite{kingma2013auto} to represent the diverse behaviors  and use HRL to train a critic network to select latent plans. Our latent-conditioned closed-loop policy can then execute the plan for obstacle avoidance and precision grasping.

%  Instead, our work focuses on the multi-modality of potential grasping plans and uses latent variables in a conditional VAE to represent the diverse behaviors. Moreover, We use HRL to train a critic network in order to select latent plans. The latent-conditioned closed-loop policy can then use the   for obstacle avoidance and precision grasping in the context of cluttered-scene grasping.
 
% \lirui{not sure if more about traditional MPC should be mentioned. IRIS or other works do not mention model-based method.s}
% Different from other deep MPC work \cite{bejjani2021learning,hirose2019deep} that use waypoints, images, or direct controls that have limited horizon, each trajectory in our method is a  grasping trajectory and is implicitly represented as a low-dimension latent vector.

 \begin{figure*}
    \centering
  \includegraphics[width=1.0\linewidth]{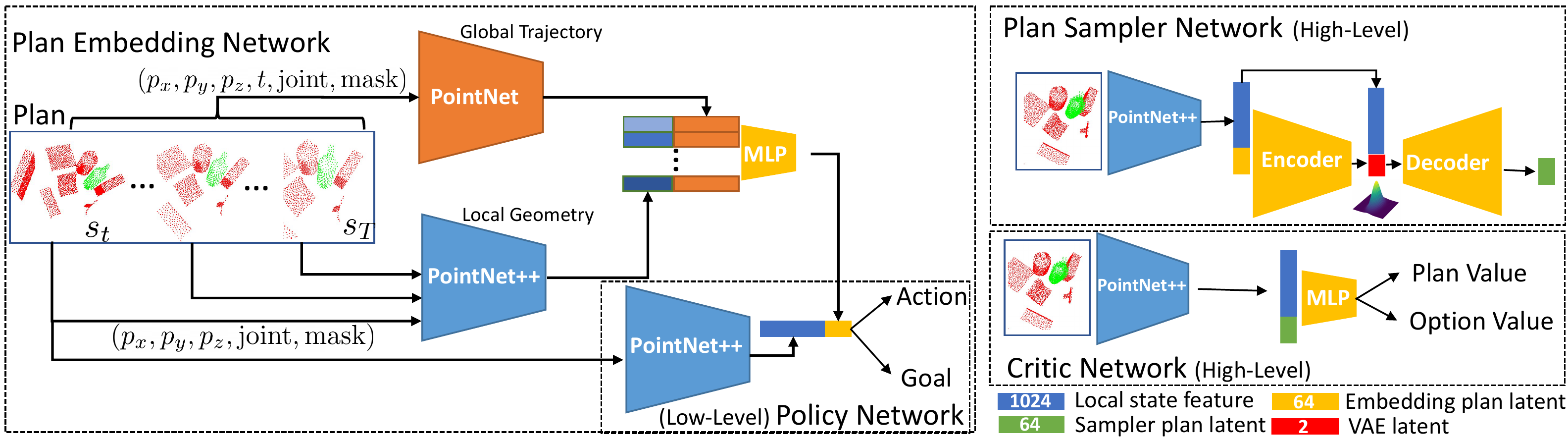}
\caption{Illustration of the network architectures {used in the hierarchical policy}. Left) Network architectures for the embedding network and low-level policy. Right) Network architectures for the VAE sampler and critic network. {All networks are trained with off-policy data except that the high-level critic network and the option classifier are trained with on-policy data.}}
\label{fig:embedding_network}
\vspace{-4mm}
\end{figure*}
%   \caption{Illustration of our offline and online training procedure. Left) During offline stage, given a noisy demonstration dataset, we jointly learn a plan embedding network with a low-level plan-conditioned policy to minimize imitation learning loss. We also train a conditional VAE that encodes and decodes the plan embedding. Right) During online stage, we use Q learning procedure by treating sampled plans as actions jointly with an GA-DDPG \cite{wang2021goal} option classifier learned from self-supervision labels.    }

\section{Learning Cluttered-Scene Grasping with Latent Plans}
\label{sec:method}

We consider the task of grasping a specific object from a cluttered scene without collisions, where observations come from a wrist camera mounted on the end-effector of a robot. To effectively solve the cluttered-scene 6D grasping problem, we use a hierarchical grasping policy consisting of two levels: the high-level policy that outputs latent plans, and the low-level policy that takes the plan and outputs pose actions at a high frequency. {The latent plan $z$ is a continuous $64$-dimension vector that encodes observations along a trajectory.} As illustrated in Fig. \ref{fig:mpc}, the system comprises of a latent plan sampler $D_\phi$, a latent plan critic $Q$, and a low-level policy $\pi$. At each planning timestep, $D_\phi$ samples $N+1$ latent plans $\{z_i\}_{i=0,...,N}$. The plan critic $Q$ scores each plan and selects the best $z_*$. At each step, the low-level policy $\pi$ takes the observation $s$ and $z_*$ to output action $a$ as the relative 3D translation and the 3D rotation of the robot end-effector.  %\xiangyun{The switching classifier is missing.} \lirui{I think it's fine to skip here, unless there is a simple way to say the low-level policy can be either pi or GA-DDPG.}

% \subsection{Target-Centric Workspace State Representation}
% \label{subsec:state}
% Following \cite{wang2021goal}, we use accumulated 3D segmented point cloud representation in the egocentric view to represent the state $s_t$. The point cloud state is augmented with a mask channel to specify the target, i.e. $0$ represents target and $1$ represents obstacles.  The point aggregation is important for ego-view agent in cluttered scene grasping whenever possible since previous observations provide important information and the wrist camera can lose views during collision avoidance.The operational space policy takes arm collisions into account by processing the robot joint information. Shown in Fig. \ref{fig:embedding_network}, we concatenate the current joint angles and remaining timestep to each point in a channel-wise fashion. This state representation improves policy learning and reduces sim-to-real perception gaps.  

\subsection{Latent Plan Embedding and Plan-Conditioned Policy}
We use accumulated segmented point clouds \cite{wang2021goal} in the egocentric view to represent the state $s_t$. In order to aggregate 3D points from different time steps, we transform the point clouds up to time $t$ to the robot base frame, and then transform the points back to the end-effector frame at time $t$. This point aggregation is important  in cluttered scene grasping since the wrist camera may not get a complete view of the scene as it moves. The point cloud is augmented with a binary mask to indicate the target, along with the current joint angles to consider arm collision avoidance. Specifically,  each point is a tuple $(p_x,p_y,p_z,q,b)$ including positions $(p_x,p_y,p_z)$, joint angles $q \in \mathbb{R}^7$, and a binary mask denoting target and obstacles $b \in \{0,1\}$. 
 
%  In our setting, the rollout $\tau$ is generated by adding perturbation noises to the expert plan to mitigate the distribution shift for standard behavior cloning \cite{ross2011reduction,laskey2017dart}.
 
To train a policy on a cluttered-scene grasping dataset with potentially many solutions (e.g. different ways to grasp a bowl), we condition the policy on a specific plan \cite{lynch2020learning} such that the low-level policy only needs to learn uni-modal behaviors. We adopt OMG planner \cite{wang2019manipulation} as the expert due to its flexibility in choosing grasping goals in cluttered scenes. We model the plan as a sequence of point clouds with respect to the end effector frame. This is computed through transforming the current observation $s_t$ by the relative poses in the trajectory waypoints. Therefore, the embedding network needs to encode both the trajectory path and the scene geometry into a low-dimension vector $z$. 
 
{In Fig. \ref{fig:embedding_network} left, the embedding network, denoted as $\theta$, encodes the point cloud trajectory {into a global plan embedding vector $z_\theta$}. 
    %   L_{z_\theta,\pi}=\frac{1}{T}\sum^T\limits_i ( L_{\text{POSE}}(g_i, g_i^*)+L_{\text{POSE}}( a_i,a_i^*))
  Specifically, let $\xi_t = (\mathcal{T}_t,  \mathcal{T}_{t+1}, \ldots, \mathcal{T}_T)$ be a trajectory of the robot gripper pose generated by the OMG planner to grasp an object at state $s_t$, where $\mathcal{T}_t \in \mathbb{S}\mathbb{E}(3)$ is the gripper pose in the robot base frame at time $t$.  Given the current state $s_t$ and the plan $\xi_t$, we represent the plan observations as a sequence of point clouds from $t+1$ to $T$ based on the poses in the plan, i.e. $\hat{s}_{t+i}=\mathcal{T}_{t+i}\mathcal{T}_{t}^{-1}(s_t)$. In Fig \ref{fig:embedding_network}, the input to the PointNet++ is each local point cloud $s_{t+i}$ and the input to the PointNet is the combined global point cloud trajectory $\cup_{\tau=t}^T \hat{s}_\tau $ with an extra normalized time channel. The encoded features are concatenated and processed through a multi-layer perceptron (MLP) as the plan embedding $z_\theta$. $z_\theta$ is then used with the current observation for the low-level policy $\pi$.   }

 {The low-level policy uses PointNet++ to extract a state feature vector from the current observation $s$ and concatenates the state feature with the latent plan $z_\theta$ as input. It outputs actions and goals which are then minimized with the behavior cloning loss and the goal auxiliary loss \cite{wang2021goal}. Let $a$ and $g$ be the action and goal predicted from the policy $\pi(s,z_\theta)$, and $a^*$ be the expert action and $g^*$ be the goal in the plan. The expert action at time $t$ is computed as $a^*_t=\mathcal{T}_{t+1}\mathcal{T}_{t}^{-1}$ which represents the relative transformation to the next timestep with respect to the current gripper pose. The expert goal at time $t$ is $g^*_t=\mathcal{T}_{T}\mathcal{T}_{t}^{-1}$ which represents the relative transform to the final timestep. For a single trajectory $\tau$, we train the networks with the supervised learning loss}
\begin{equation}
\label{embedding}
       L_{\text{traj}}(z_\theta,\pi)=\frac{1}{T}\sum^T\limits_i ( L_{\text{pose}}(g_i, g_i^*)+L_{\text{pose}}( a_i,a_i^*)).
\end{equation}
 $L_{\text{pose}}$ \cite{wang2021goal} is the point matching loss for a set of predefined gripper points to optimize translation and rotation:
\begin{equation} \label{eq:pose}
    L_{\text{pose}}(\mathcal{T}_1, \mathcal{T}_2) =\frac{1}{|X_g|}\sum\limits_{x\in X_g} \| \mathcal{T}_1(x)-\mathcal{T}_2(x) \|_1,
\end{equation} 
where $\mathcal{T}_1, \mathcal{T}_2 \in \mathbb{S}\mathbb{E}(3)$, $X_g$ denotes a set of pre-defined 3D points on the robot gripper, and $L_1$ norm is used. The same pose loss applies to all actions and goal predictions. {Similar to \cite{wang2021goal}, we add the goal auxiliary loss to behavior cloning to align the policy intent with the encoded demonstrations and improve network interpretability and training.}

Minimizing the trajectory loss jointly enables a shared latent plan representation between the policy network and the latent embedding network. Then the high-level policy can sample and evaluate the latent plans as the action space and the low-level policy can act as a closed-loop controller to execute the latent plan. On Fig. \ref{fig:latent_vis} left, we plot the 2-dimension t-SNE embeddings for the dataset, and denote the plan manifold as $\mathcal{Z}$. We observe that $\theta$ implicitly learns to distinguish plan embeddings from different trajectories and cluster embeddings for states in the same trajectory.

% The plan is encoded into $64$ dimension vector with spatiotemporal network \cite{rempe2020caspr} to capture the trajectory and geometry details. The low-level policy concatenates current observation and plan latent to output action and goal. The critic and sampler architecture uses Pointnet++ \cite{qi2017pointnet++} and Multi Layer Perceptron as the main architectures.
% \begin{figure*}
% \centering 
% \includegraphics[width=0.9\textwidth ]{img/rollout_vs_plan_vs_recons.png}
%   \caption{ Plan v.s. rollout and reproducing}
%   \label{fig:contacts}
% \end{figure*}

% \lirui{Reduce problem complexity to the nominal planner represented as a generative model. Nominal plan, behavior, high-level action, motion primitive, motion sketch, latent skill, trajectory, all refer to the same thing. We need to settle down the notations. }
%  Since the embedding $z$ is low dimension and captures abstract plan information, the latent $c$ is $2$ dimension in our implementation. The problem of combined obstacle avoidance and instance grasping in cluttered scenes is then transformed into a planning problem in the low-dimensional trajectory embedding space $\mathcal{Z}$.

\subsection{Latent Plan Proposal with a Generative Model}
\begin{figure*}
\centering 
\includegraphics[width=0.65\linewidth ]{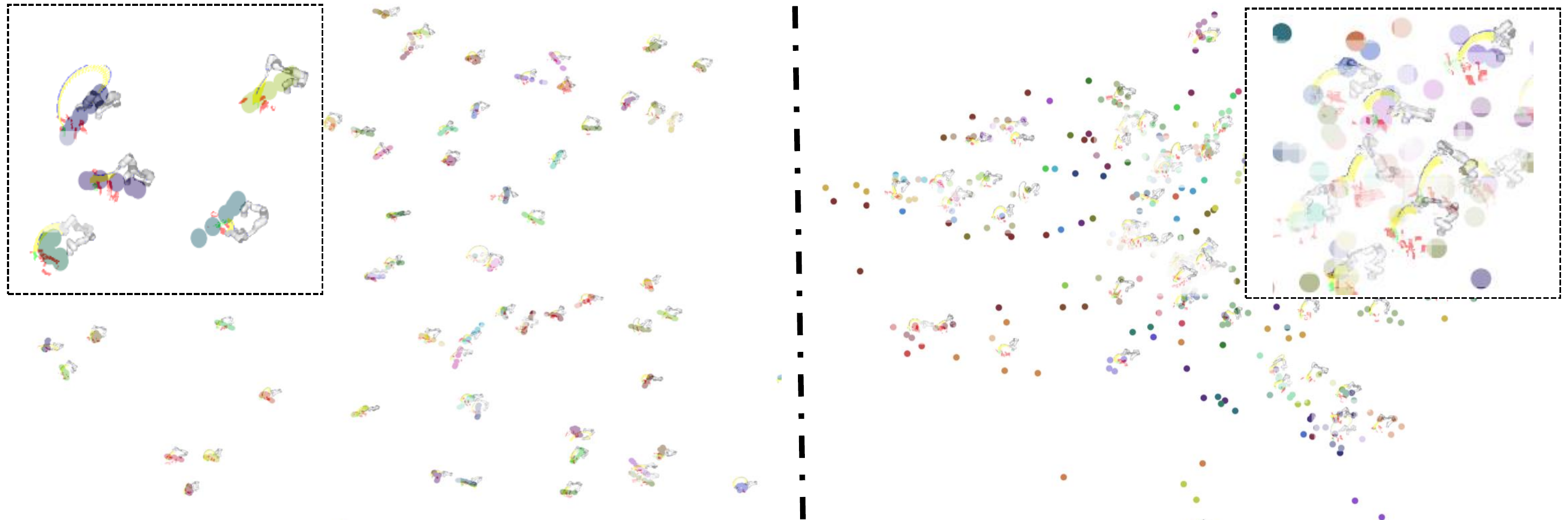}
  \caption{Comparison of t-SNE embeddings and VAE embeddings of latent plans $z_\theta$ where states in the same trajectory are denoted by the same color. Left) The 2-dimension t-SNE embedding of $\mathcal{Z}$. Right) The 2-dimension VAE latent $\mathcal{C}$.  }
  \label{fig:latent_vis}
  \vspace{-4mm}
\end{figure*}

To generate latent plans without experts for new scenes, we propose to use a variational inference framework~\cite{kingma2013auto}. The objective of the latent plan sampler is to model the distribution of state observations $s$ and the expert plan embedding $z$, to provide multi-modal plans at inference time. We use a conditional variational autoencoder (cVAE), denoted as $\phi$, to learn this distribution. A cVAE includes a parametrized encoder $E_\phi$ and decoder $D_\phi$, which we model as MLPs (Fig. \ref{fig:embedding_network} top right). During training, we maximize the marginal log likelihood of the trajectory embeddings in the dataset $\log_\phi(z_\theta|s)$. At test time, we sample from a unit Gaussian to generate the expert plan embedding. The Evidence Lower Bound (ELBO) \cite{kingma2013auto} for the observation-conditioned VAE can be written as 
\begin{equation}
\label{kl}
\log p_\phi(z|s)\ge -\mathbf{KL}(q(c|z,s))\vert \vert  p(c))+\EXP0_{q(c|s,z)}[\log p(z|c,s)],
\end{equation} 
where $c=E_\phi(s,z)$ denotes the VAE latent variables encoded from the plan embedding $z$ and conditioned on the state $s$. The state feature is extracted with a PointNet++ and concatenated with the latent plan (Fig. \ref{fig:embedding_network}). The prior distribution $p(c)$ is restricted to be a 2-dimension unit Gaussian. In Fig. \ref{fig:latent_vis} right, we visualize the 2D learned VAE latent space    $\mathcal{C}$. 

% The VAE learning objective balances between representing salient plan semantics and regularizing with the gaussian prior.

To train the cVAE model, we maximize the ELBO with a Kullback–Leibler divergence regularization loss \mbox{$L_{\text{KL}}=\mathbf{KL}(\mathcal{N}(\mu_\phi, \mathbf{diag}({\sigma^2_\phi}))\vert \vert  \mathcal{N}(\mathbf{0},I_2))$} and a latent reconstruction loss $L_{\text{recons}}=\norm{z_\theta-z_\phi}^2_2$, where we parametrize the encoder with mean $\mu_\phi$ and diagonal covariance $\sigma_\phi$ and use the reparametrization tricks \cite{kingma2013auto}  to sample $z_\phi$. Additionally, the sampler latent $z_\phi$ can be further decoded with the low-level policy $\pi(s,z_\phi)$ on the expert trajectory to minimize the plan loss in Eq. \eqref{embedding}. In this way, the action and goal losses from $L_{\text{traj}}(z_\phi,\pi)$ also provide gradient information such that the latent generated by $\phi$ acts similarly to the expert plan latent $z_\theta$. Therefore, the total loss is 
\begin{equation}
\label{sampler}
    L_{\text{sampler}}=\beta L_{\text{KL}}+L_{\text{recons}}+L_{\text{traj}}(z_\phi,\pi),
\end{equation}
where $\beta=0.02$ is a hyper-parameter \cite{higgins2016beta}. The cVAE and the embedding network are trained jointly (see Algorithm \ref{Algorithm1}). At test time, we sample $c\sim\mathcal{C}$ from a unit Gaussian and decode the plan embedding $z=D_\phi(s,c)$. We observe that planning diverse trajectories with a generative model improves local behaviors such as obstacle avoidance.

  \begin{algorithm}
 \caption{Off-Policy Training (demonstration $D_{\text{offline}}$)}
\label{Algorithm1}
\For{ $i = 0,...,N$}{
    Initialize low-level policy $\pi$, plan embedding network $\theta$, and plan sampler $\phi$ \\
    Sample a rollout $\tau_t=\{(s_t,a_t,g_t),...,(s_T,a_T,g_T)\}$ and plan $\xi_t$ from $D_{\text{offline}}$ \\
    \textbf{Encode Plan:} Encode $\xi_t$ into $z_\theta$  and optimize $\pi,\theta$ on $\tau_t$ with Eq. \eqref{embedding} \\
    \textbf{Generate Plan:} Autoencode $z_\theta$ into $z_\phi$ and optimize $\phi$ on $\tau_t$ with Eq. \eqref{sampler} \\
 } 
\end{algorithm}

%  Left) The 2-dimension t-SNE embedding of $\mathcal{Z}$ shows that our embedding network $\theta$ implicitly learns to distinguish plan embeddings from different trajectories and cluster embeddings for states in the same trajectory. Right) The 2-dimension VAE latent $\mathcal{C}$ shows that VAE learning objective balances between representing salient plan features and regularizing with the gaussian prior.  

% Comparison of TSNE embedding (left) and cVAE (right) embedding plots of the latent plan space $\mathcal{Z}$. Latent embedding of sampled states for each trajectory are denoted by the same color. 

\subsection{Hierarchical RL for Latent Plan Critic}

To continually learn and improve on unknown objects and scenes, we extend the  system with hierarchical reinforcement learning.  The low-level policy, plan sampler, and plan embedding networks are fixed during on-policy learning. The trained low-level policy and sampler can help avoid the risks and expenses of random explorations. The high-level actions are the latent plans $z_\phi$ sampled from $D_\phi$.  The goal is to learn plan selection through online experience in avoiding obstacles and attempting grasps. 

Specifically, we use the standard notation of a Markov Decision Process (MDP): $\mathcal{M}=\{\mathcal{S},\mathcal{R},\mathcal{A},\mathcal{P},\gamma \}$. $\mathcal{S}$, $\mathcal{A}$, and $\mathcal{O}$ represent the state, action, and observation space. \mbox{$\mathcal{R}:\mathcal{S} \times \mathcal{A}\to \mathbb{R}$} is the reward function.   
\mbox{$\mathcal{P}: \mathcal{S}\times \mathcal{A}\to \mathcal{S}$} is the transition dynamics. $\gamma=[0,1)$ is a discount factor.  To simplify the setting, we still use $s$ to represent the policy input even though observations are used.  Our goal is to learn a policy that maximizes the expected cumulative rewards $\EXP0_{\pi}[\sum_{t=0}^\infty \gamma^t r_t]$, where $r_t$ is the reward at time $t$.

%  The Q-function of the policy for a state-action pair is $Q(s,a) =\mathcal{R}(s,a)+\gamma\EXP0_{s',  \pi}[\sum_{t=0}^\infty \gamma^tr_t|s_0=s']$, where $s'$ represents the next state of taking action $a$ in state $s$ according to the transition dynamics.
% To improve the grasping success, we adopt a simple instantiation of the options framework \cite{sutton1999between}. 

We use off-policy Q-learning \cite{van2016deep} to train the plan critic $Q(s_t, z)$ to predict the value of starting from current state $s_t$ and following the plan $z$.  % Instead of using the low-level actions $Q(s_t, a_t)$, we use the latent vector $z$ to provide information of the entire trajectory.  
We use sparse rewards for the MDP: an episode terminates  with reward $1$ if a grasp succeeds, with reward $-1$ if a collision happens, or $0$ in other cases. We use a mixture of return and bootstrapped next-state value as the target for the Bellman equation 
\begin{equation}
\label{eq:critic}
  y_t=\lambda(r(s_t,z_t)+\gamma \max_{z'\sim \phi(s_{t+1})} Q(s_{t+1},z')) +(1-\lambda )\gamma^{T-t}r_T, 
\end{equation}
 where $\lambda=0.5$. 
The training loss for the critic is \mbox{$L_{\text{critic}}=(Q(s_t,z_t)-y_t)^2$}. 
For every $12$ environment steps at test time, we resample $8$ trajectory candidates for the critic to select an optimal plan and filter out the suboptimal ones. 
 
% Since the problem reward is sparse and there is only a single option transition, the supervised learning can also be treated as option Q-learning.

% \begin{algorithm}
%  \caption{Plan Critic Training (low-level policy $\pi$, plan sampler $\phi$)}
% \label{Algorithm2}
% \For{ $i = 0,...,N$}{
%     Initialize  critic $Q$, GA-DDPG critic $G$, dataset $D_{\text{online}}$, and switch step $t_{\text{switch}}$ \\
%     Execute $\pi$ with sampled latent $z$ for $t_{\text{switch}}$ steps and switch to GA-DDPG policy \\
%     Log trajectory $\tau=(s_0,a_0,r_0,z_0...,s_T,a_T,r_T,z_T|t_{\text{switch}})$ to $D_{\text{online}}$
%       \\
%     Sample $(s_t,z_t,s_{t+1},r_T)$ from $D_{\text{online}}$ \\
%     Optimize $Q,G$ with Eq. \eqref{critic} \\
%  } 
% \end{algorithm}
\subsection{Low-level Policy Switch Option Classifier}
\begin{figure*}
\centering 
\includegraphics[width=0.85\textwidth ]{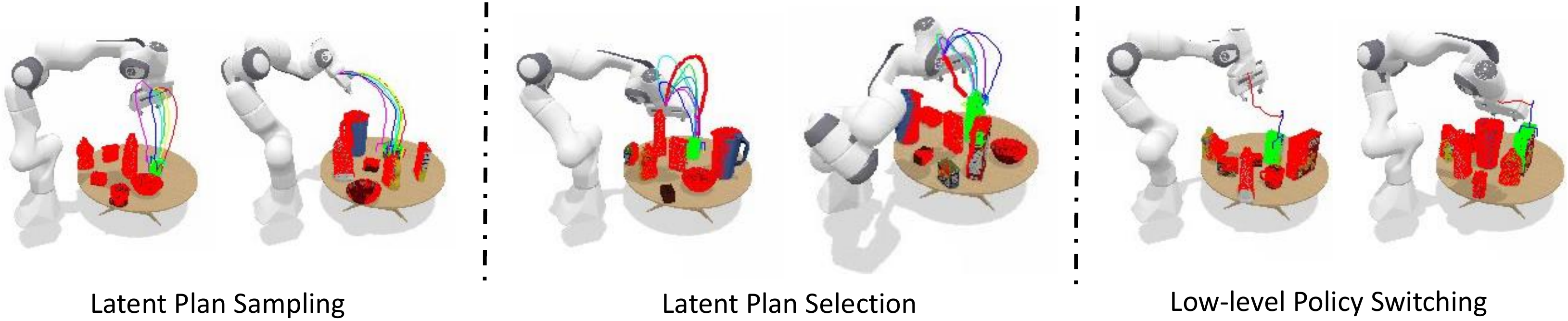}
  \caption{Qualitative analysis of the method components. Left) We observe some very distinct plans generated from the plan sampler network. We smoothly modify one latent coordinate and observe the corresponding plan transitions, colored from red to magenta in the HSV space. Middle) The latent plan critic selects the maximum value trajectory (colored in red) among samples. Right)  The policy transition from a plan trajectory (red) to a target-specific grasping trajectory (blue). }
  \label{fig:example}
  \vspace{-3mm}
\end{figure*}

  In the full system, we learn an additional option classifier $G(s_t)$ to choose the binary policy option. It allows to switch the low-level policy from $\pi$ to a pretrained GA-DDPG policy \cite{wang2021goal}, denoted as $\pi_g$, to execute workspace control $a$. GA-DDPG is trained to focus on grasping single objects. Intuitively, once the robot avoids the obstacles, using GA-DDPG as a single-object grasping primitive is more accurate for final grasping. $G(s_t)$ learns to predict the episode success if we start using $\pi_g$ from the current state $s_t$. We learn the GA-DDPG option classifier $G(s_t)$ jointly with the critic $Q(s_t, z)$ (Fig.~\ref{fig:embedding_network}) by augmenting GA-DDPG execution from a random timestep $t_{\text{switch}}$.   The final episode reward can be used as the label for $G(s_t)$ in timesteps after $t_{\text{switch}}$. The loss for the option classifier is
  \begin{equation}
  \label{eq:option}
   L_{\text{option}}=\mathbf{BCE}(G(s_t), r_T^{+})\cdot \mathds{1}(t\ge t_\text{switch}),  
  \end{equation}
  where $\mathbf{BCE}$ denotes the binary cross entropy loss and $r_T^{+}=1$ if  $r_T=1$ and $0$ otherwise.  The on-policy training algorithm is summarized in Algorithm \ref{Algorithm2}. Examples of plan sampling, plan selection, and policy switching are shown in Fig. \ref{fig:example}.

\begin{algorithm}[htb]
 \caption{On-Policy Training (plan sampler $\phi$, low-level policy $\pi$, replay buffer $D_{\text{online}}$, GA-DDPG pretrained policy $\pi_g$)}
\label{Algorithm2}
\For{ $i = 0,...,N$}{
    Initialize  critic $Q$, policy option classifier $G$, and switch step $t_{\text{switch}}$ \\
    Execute $\pi$ with sampled latent $z$ for $t_{\text{switch}}$ steps and switch to  $\pi_g$ to execute grasping \\
    Log trajectory $\tau=(s_0,a_0,r_0,z_0,...,s_T,a_T,r_T,z_T)$ to $D_{\text{online}}$  \\
    Sample a transition $(s_t,z_t,r_t,s_{t+1})$ with episode information $(r_T,t_{\text{switch}})$ from $D_{\text{online}}$ \\
    \textbf{Select Plan and Policy:} Optimize $Q$ with Eq. \ref{eq:critic} and $G$ with Eq. \ref{eq:option} \\
 } 
\end{algorithm}
 \begin{section}{Implementation Details}
 \label{appendix:impl}
  \begin{subsection}{Task and Environment}
We experiment with the Franka Emika Panda arm, a $7$-DOF arm with a parallel gripper. We use ShapeNet \cite{chang2015shapenet} and YCB objects \cite{calli2015benchmarking} as the object repository and the Acronym \cite{acronym2020} grasp set for grasp planning. A task scene is generated by first placing a target object, and then placing up to $7$ obstacles with random stable poses on a table in the PyBullet simulator \cite{coumans2013bullet}. We use a maximum horizon of $T=50$ and use sparse rewards for the MDP. An episode succeeds with reward $1$, collides with reward $-1$, or $0$ in other cases. We measure collisions with the built-in contact detectors in Bullet. We use approximately $1,500$ ShapeNet objects from $169$ different classes for offline training and YCB scenes for online interactions. For testing, we select $9$ YCB objects with 20 unseen scenes per object. We test in each scene for 5 times and compute the statistics of grasping success rates, collision rates, and rewards.  

At every timestep, the observation has color, depth, and mask images of size $224\times 224$. We backproject the depth images with foreground masks to generate the scene point clouds. When adding new observed points to the global aggregated point cloud, we use farthest point sampling \cite{qi2017pointnet++} to limit the number of total points. The input to the network contains $4096$ sampled points ($1024$ target points and $3072$ obstacle points).  % For the action parametrization, we adopt workspace control for its alignments with the observation space and learning performance. 
We randomize the initial states $\rho_0$ from a uniform pose distribution facing toward the target object, with a distance ranging from $0.6$~m to $0.8$~m. When points are detected between the fingertips, the robot will close its gripper and retract the arm to complete a grasp.  
 
%  We have experimented with adding sampled robot arm point clouds and table point clouds, as well as computing nearest collision point distances and object point normals as inputs, and did not observe performance improvements. Instead of modeling collision-specific features or designing analytic cost functions, the high-level latent critic learns collision-avoidance from the sparse collision rewards through interactions.
 \end{subsection}

 \begin{table*}
\centering
\begin{minipage}{0.8\linewidth}
\resizebox{\linewidth}{!}
{\begin{tabular}{|c|cccccc|}
\hline
\multicolumn{0}{|c|}{Test}   &
  \multicolumn{0}{c}{BC} &  \multicolumn{0}{c}{GA-DDPG} & \multicolumn{0}{c}{BC Residual} & \multicolumn{0}{c}{Waypoint} & \multicolumn{0}{c}{Rollout Encoding}    & \multicolumn{0}{|c|}{Plan Encoding}    \\ \hline
Reward & 0.128 +- 0.03      &  -0.240 +- 0.05  & 0.273 +- 0.02  & 0.044 +- 0.02 &0.463 +- 0.04   & \textbf{0.600 +- 0.03}  \\
Success &  0.437 +- 0.02     & 0.328 +- 0.02 &   0.463 +- 0.01  &  0.356 +- 0.03  & 0.618 +- 0.03  &   \textbf{0.705 +- 0.02}  \\
Collision &   0.308 +- 0.03   & 0.570 +- 0.03   & 0.190 +- 0.02    & 0.312 +- 0.02  &   0.157 +- 0.01 &   \textbf{0.104 +- 0.01}  \\
\hline
\end{tabular}}
\end{minipage}
\caption{Evaluation statistics (one standard deviations) of different policy methods. }
\label{table:exp1}
\vspace{-4mm}
\end{table*}
% prev latent, visualization etc
\begin{table*}
\centering
\begin{minipage}{0.8\linewidth}
\resizebox{\linewidth}{!}
{\begin{tabular}{|c|cccccc|}
\hline
\multicolumn{0}{|c|}{Test}   &
  \multicolumn{0}{c}{Regress }   & \multicolumn{0}{c}{Vanilla VAE} & \multicolumn{0}{c}{Zeros}  & \multicolumn{0}{c}{No Policy Loss}  & \multicolumn{0}{c}{Single Plan} & \multicolumn{0}{|c|}{Plan VAE}   \\ \hline
Reward &  0.369 +- 0.01  &  -0.460 +- 0.02 & 0.385 +- 0.04   & 0.400 +- 0.05  & 0.492 +- 0.02 &  \textbf{0.536 +- 0.03}   \\
Success & 0.600 +- 0.01  &  0.022 +- 0.01 & 0.543 +- 0.02   & 0.528 +- 0.05 &  0.601 +- 0.02  &  \textbf{0.637 +- 0.02}  \\
Collision &  0.231 +- 0.01  & 0.483 +- 0.02 & 0.158 +- 0.01  &   0.126 +- 0.01 & 0.110 +- 0.01 & \textbf{0.101 +- 0.01}    \\
\hline
\end{tabular}}
\end{minipage}
\caption{ Evaluation statistics of different models that are used to generate the trajectory latent.}
\label{table:exp2}
\vspace{-4mm}
\end{table*}

 \begin{subsection}{Network Architecture}
  \label{appendix:network}
 Shown in Fig. \ref{fig:embedding_network}, our plan embedding network is similar to \cite{rempe2020caspr}. PointNet++ \cite{qi2017pointnet++} is used to capture the local geometry and PointNet \cite{qi2017pointnet} is used to extract the global plan information. The variable-length trajectory  features for each point cloud are then maxpooled into a global feature $z_\theta$ of dimension $64$. For the trajectory latent autoencoder, we use a $1024$-dimension  state features extracted by PointNet++. The critic, the sampler, and the policy network each uses a separate PointNet++ with $1024$ dimension features to represent state $s$.  The  latent variable $c$ in the cVAE space $\mathcal{C}$ has dimension $2$ and is decoded into $64$ dimension $z_\phi$. We use group norm instead of batch norm since the states in a trajectory are highly correlated. To provide information about the horizon to the network, the remaining timestep $T-t$ is concatenated for all networks.  The actor network and the critic network are both implemented as 3-layer MLPs with ReLU activation. We use Adam \cite{kingma2014adam} optimizer to train all the networks. 
 
\end{subsection} 

\begin{subsection}{Training Details}
 \label{appendix:train_test}
Our offline training phase uses $120000$ demonstration data points (around $2400$ trajectories) from random scenes with ShapeNet objects. 
  To fit into the GPU memory for training, a point cloud trajectory is evenly downsampled with ratio $6$ and the supervised trajectory states are randomly downsampled with ratio $5$. We include both end points for trajectory encoding and supervision, so the length of a downsampled trajectory varies from 2 to 11 for a horizon of $T=50$. To improve the robustness to imperfect actuation, the rollout $\tau$ is generated by executing a perturbed version of the expert plan, and the supervision is obtained by running OMG-Planner on the perturbed trajectory. This helps mitigate the distribution shift problem in behavior cloning.
  
During the HRL phase, we jointly train the critic and the switching classifier, whereas the sampler network, the plan embedding network, and the low-level policy are frozen. We sample $18$ parallel actor rollouts to collect experiences and  perform $50$ training optimization steps for each episode. We resample $8$ trajectory latents with random probability $0.1$. GA-DDPG policy switching is added with random steps and increases over time. The MDP discount factor $\gamma$ is set to $0.99$.

% We have experimented with the common RL fine-tuning approach of training sampler and critic jointly with mixed offline and online data and observe that the multi-modal plan generation is affected.  We have experimented with other generative models and find no better performance.We want to keep diverse planning to handle the multi-modality for grasping in clutter and potentially incorporate other scoring methods.  
 \end{subsection}

 \end{section}

\section{Experiments}

\label{sec:experiment}
We conduct experiments and present our findings to the following questions: (1) Given an expert plan, how well do the trajectory encoding network and the plan-conditioned policy behave compared with other end-to-end baselines? (2) Does the cVAE generative model learn diverse and feasible planning behaviors for unseen test scenes? (3) Can the latent critic and GA-DDPG policy classifier trained with HRL further improve the performance? Finally, we also conduct comparisons with open-loop baselines in simulation and evaluate our system in the real world.

\subsection{Plan Embedding and Trajectory Execution}
We conduct experiments to investigate the embedding representation and the low-level policy. {Note that several baseline methods do not use the extra embedding information.} In Table~\ref{table:exp1}, we can see that i) End-to-end behavior cloning (``BC'') does not perform well due to the multi-modal trajectory demonstrations. The state space for grasping in clutters is also more complex than single object grasping. ii) Directly running pre-trained GA-DDPG (``GA-DDPG'') policy without considering the obstacles results in frequent collisions. One common extension is to learn a residual policy (``BC-Residual'') \cite{silver2018residual} on top of GA-DDPG by summing up the action predictions. This also performs suboptimally. iii) A vanilla planning network that outputs trajectory waypoints (``Waypoint'') behaves similarly to BC. It shows that representing plans with latent variables compatible with a closed-loop policy is better than an extended action space. The closed-loop observation-conditioned controllers are more suitable for executing the long-horizon rollout. iv) The model that encodes a noisy and dynamics-involved rollout (``Rollout Encoding'') performs worse than plan encoding. Overall, the path geometry and local observation information in the latent plan enables the plan-conditioned policy to outperform naive end-to-end baselines. 

% Furthermore, we observe that the low-level policy learns reactive motions to avoid obstacles.

% & \multicolumn{0}{c|}{Plan + GA-DDPG}
\subsection{Generative Models for Latent Plans}
With the same low-level policy and embedding network, we evaluate the performance of different variations of the latent planner. In Table~\ref{table:exp2}, we can see that i) Latent regression (``Regress'') models have trouble providing useful plan information in novel test scenes. A cVAE models the joint state and latent distribution and does better at capturing the multi-modality in the plan space. We also observe that models trained without a separate latent VAE network \cite{lynch2020learning} tend to collapse the plans during inference. ii) A vanilla $64$ dimension latent VAE (``Vanilla VAE'' trained with no hierarchy on $z$) only provides a unit Gaussian sample to the low-level policy and is  insufficient to represent useful behaviors. iii) A dummy zero latent (``Zeros'') also does not provide as much information to the low-level policy. iv) Compared to methods that only reconstruct the latents (``No Policy Loss''), additional policy loss improves training the latent model. v) Reactively sampling new plans in a closed loop is better than adopting a single plan (``Single Plan'') for the entire execution. Finally, inferring latent plans without experts still outperforms end-to-end policies. 

% Since the low-level policy is trained to map from the observation and latent to goals and actions, plan latent contains additional information to curve around obstacles and to reach diverse goals. When the latent is not meaningful, the policy often just follows a straight trajectory to the nearest goal, which is prone to colliding with other objects.

\begin{figure}
 \centering
\begin{subfigure}{0.5\textwidth}
  \includegraphics[width=\linewidth]{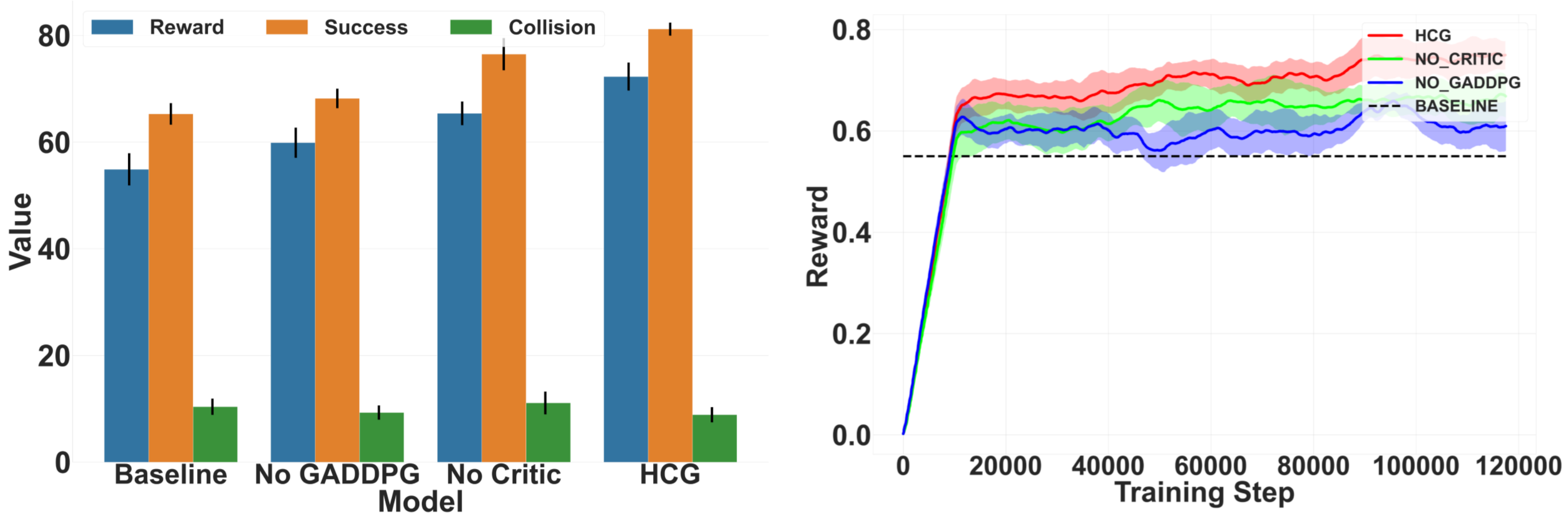}
\end{subfigure}
% \begin{subfigure}{0.23\textwidth}
%   \includegraphics[width=\linewidth]{img/rew_plot.pdf}
% \end{subfigure}
\caption{Left) Statistics for the reward plots. Right) Ablation study for the on-policy RL training stage.  We observe that training plan selection and GA-DDPG option switch improves the policy.}
\label{fig:rl_rew_plot}
\vspace{-3mm}
\end{figure}

\subsection{Hierarchical Reinforcement Learning}
 \begin{figure*}
 \centering 
 \includegraphics[width=0.6\linewidth ]{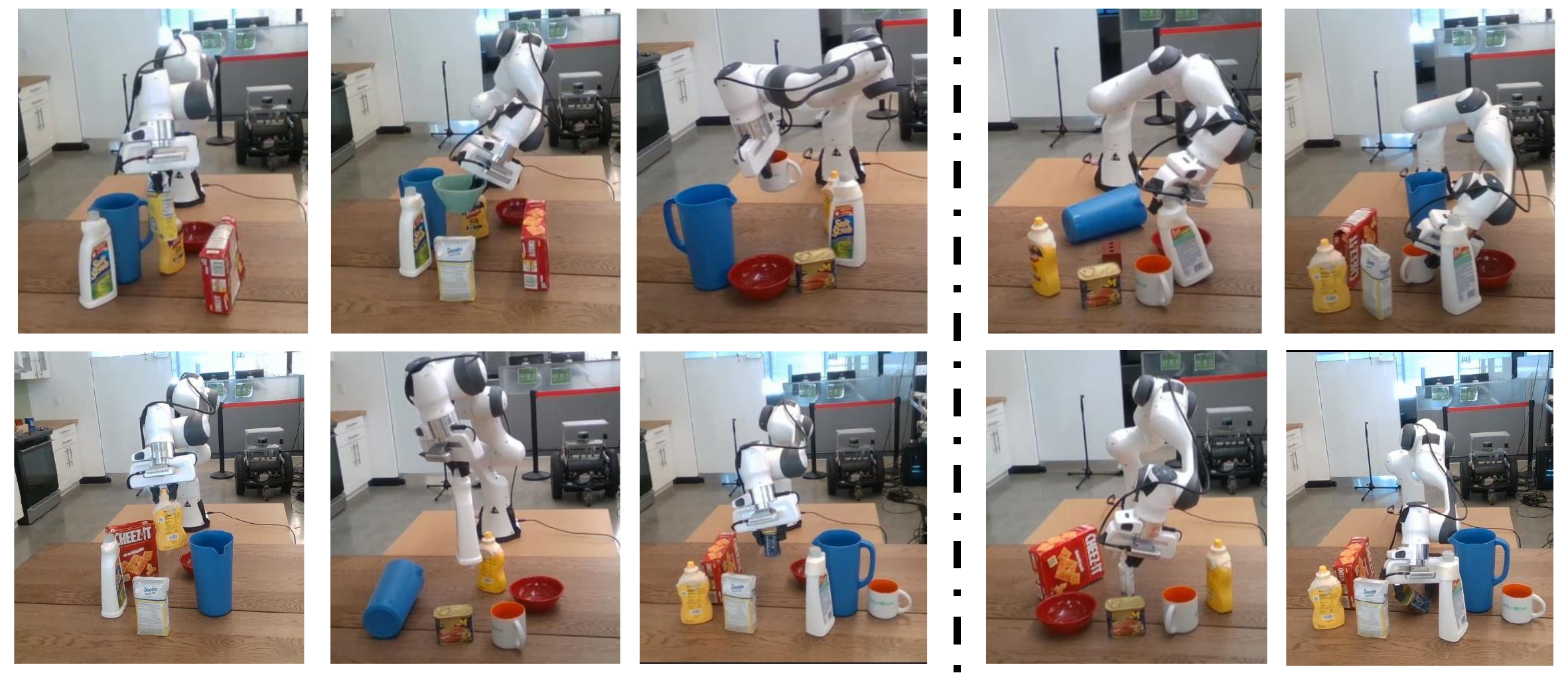}
   \caption{Successes (left) and failures (right) of real-world tabletop cluttered-scene grasping experiments with
our policy trained in simulation. The target object is surrounded by obstacles and  the robot initial configuration is different for each scene.}
   \label{fig:real_world_success}
   \vspace{-3mm}
\end{figure*}
\begin{table*}
\small
\centering
\begin{minipage}{0.8\linewidth}
\resizebox{\linewidth}{!}
{\begin{tabular}{|c|ccccc|}
\hline
\multicolumn{0}{|c|}{Test}   &
  \multicolumn{0}{c}{6DGraspNet }   & \multicolumn{0}{c}{Contact-GraspNet} & \multicolumn{0}{c}{Multi-view 6DGraspNet} & \multicolumn{0}{c}{Closed-loop 6DGraspNet}  & \multicolumn{0}{c|}{HCG}   \\ \hline
Success & 0.67  &  0.72 & 0.78 & 0.75 &  \textbf{0.82}  \\
Collision &  0.19  & 0.16 & 0.12  & 0.10 & \textbf{0.09}    \\
Inference Time &  1.42  & 0.28 &  1.67 &  1.42  & \textbf{0.04}    \\
Execution Time &  31.21 & 29.55 & 36.63 &  52.88 & \textbf{21.14}   \\
\hline
\end{tabular}}
\end{minipage}
\caption{Performance comparisons with the different baseline methods. Inference time only considers the network forward pass while the execution time includes grasp generation, motion planning, and the actual execution.}
\label{table:baseline_computation}
\vspace{-4mm}
\end{table*}
\begin{table}

\centering
\begin{minipage}[t]{.9\linewidth}
\resizebox{\linewidth}{!}
{\begin{tabular}{|c|ccccc|}
\hline
\multicolumn{0}{|c|}{Obstacle Number}   &
  \multicolumn{0}{c}{3} & \multicolumn{0}{c}{4} & \multicolumn{0}{c}{5} & \multicolumn{0}{c}{6} & \multicolumn{0}{c|}{7} \\ \hline
HCG Success & 0.90  & 0.84   & 0.78  &  0.75  & 0.73   \\
Open-Loop Method Success & 0.89 & 0.86 & 0.75 & 0.68 & 0.64   \\
\hline
\end{tabular}}
\end{minipage}
\caption{{Policy success rates when varying the level of clutter. More obstacles increases the difficulty.}}
\vspace{-3mm}
\label{tab:clutter}

\end{table}

Our policy trained on offline demonstration data and ShapeNet objects achieves a mean reward of $0.54$ (Table \ref{table:exp2}). Compared to pure learning from demonstrations, our hierarchical RL trains the critic and the GA-DDPG option classifier on new scenes with interactions.  Shown in Fig.~\ref{fig:rl_rew_plot}, we observe an improvement in the final reward and plan selection in addition to the combination of sampler and low-level policy. RL training enforces the critic to learn a useful scoring function that improves stability and collision in hard scenes. Moreover, GA-DDPG policy helps align the grasping action to improve the final grasping accuracy. In test scenes, our final hierarchical policy reaches $81.2\%$ success rate with only $8.9\%$ collision. {We also ablate on the policy performance for different levels of clutters in Table \ref{tab:clutter}. Due to higher occlusions and more difficult obstacle avoidance, the policy performance is affected by the clutter levels. These scenes are even more challenging for open-loop traditional methods since the initial scene observations do not have sufficient information for planning.}

%The plans can be further combined with analytic trajectory scoring methods or trajectory optimizers before execution. 
 
%  The plan selection improves stability and collision in hard scene instances where the arms need to carefully avoid the obstacles. 
 
% 

% \begin{figure}
%     \centering
%   \includegraphics[width=0.9\linewidth]{img/rl_rew_plot.pdf}
% \caption{Left) Test performance for ablation on the self-supervision stage. Right) The mean and the $1$ standard deviation for the reward plots. We observe that the combination of plan selection and GA-DDPG option improves the performance.}
% \label{fig:rl_rew_plot}
% \vspace{-2mm}
% \end{figure}

% The open-loop method with 6DGraspNet achieves $67.2\%$ success rate with $19.0\%$ collisions and the one with Contact-Graspnet achieves $72.2\%$ success rate with $16.4\%$ collision rates due to better grasps.  The forward pass of the grasp detection itself in open loop methods runs at 0.7 fps. And contact graspnet runs at 0.28s per grasp detection. 

\subsection{Comparison with Open-Loop Baselines}
We compare our method against the state-of-the-art grasp detection method 6DGraspNet \cite{mousavian20196,murali20196} and its variant Contact-Graspnet \cite{sundermeyer2021contact} (designed to work in cluttered scenes) in simulation. The manipulation pipeline approximates the scene SDF with point clouds and runs grasp detection methods to generate diverse grasps as the goal set. The OMG planner \cite{wang2019manipulation} uses the goal set to select a grasp and generate a joint trajectory. When the planner returns no solution due to bad grasp sampling, we still execute a failed plan for fair comparison. We also introduce two additional baselines: ``Close-loop 6DGraspNet'' runs grasp detection and planning every 5 environment steps with the aggregated point clouds. It has fewer collisions and better grasping performance due to the availability of additional state information.  ``Multi-view 6DGraspNet'' collects 3 sample views of the scene point clouds at the beginning of the episode to have a more complete point cloud representation. In Table \ref{table:baseline_computation}, we observe that our method outperforms these baselines and is much faster. Note that the test scenes contain a variety of occlusions and obstacles, so closed-loop perception and control are required for collision-free grasping. With only a single view, a robot cannot see what is behind an object due to occlusions, which results in grasps that are not aligned well with the shape geometry. This shows that closed-loop perception with a hierarchical planning can substantially improve grasping success rates in clutters. % Our method also  requires less computation time. % and hand-tuning for specific scenarios.
% We leave comparisons with close-loop manipulation baselines to future works.

%   than many real-world test scenes where objects are packed in rows and collisions are unlikely to happen Since many failures are due to collisions during motions, incorporating a collision detection network \cite{danielczuk2020object} into a framework open-loop methods could improve the performance. 

\subsection{Tabletop Cluttered-Scene Grasping in the Real World}
We conduct real-world grasping experiments with HCG. To obtain the segmented object point clouds in the tabletop setting, we use an unseen instance segmentation method~\cite{xiang2020learning} that runs at 5 fps. At inference, the latent plan sampling and selection take around $0.2$s and the low-level policy runs at $25$ Hz. This is considerably faster than traditional methods that require expensive grasp sampling and  motion planning. Fig. \ref{fig:real_world_success} shows some qualitative results in our experiments where the target object is surrounded by obstacles. Compared to single-object grasping or top-down picking in a piles, the experiment setup is challenging with heavy occlusions and requires intricate collision avoidance. As in simulation, the robot grasped each target object 5 times with varying initial configurations. Our model trained only in simulation generalizes well to the real world. Our model generates plans to avoid arm collisions and produces smooth controls to grasp objects in these challenging cluttered scenes. Overall, our method succeeds with $21$ trials out of $34$ attempts on $8$ challenging test scenes. The majority of failures is due to slippery contacts caused by sim-to-real physics gaps and small collisions caused by suboptimal plans and controls. One potential way to improve this is to train on more diverse data in simulation and finetune the policy in the real world.

\section{Conclusion}

\label{sec:conclusion}
We introduce Hierarchical policies for Cluttered-scene Grasping (HCG),  an end-to-end method to grasp in clutters from point clouds. Our method learns a latent plan sampler through variational autoencoder, a latent plan critic through Q learning, and an option classifier to improve the grasping accuracy. The overall system achieves robustness and generalization across diverse unseen scenes and robot initial configurations both in simulation and in the real world. Future directions include incorporating other modalities for grasping and lifting the segmentation requirement for the policy.

% The task is common in unstructured human environments and yet is challenging for current policy methods and traditional manipulation pipelines due to the complexity in partial views and collision avoidance.  
 
% \bibliographystyle{IEEEtran}
\bibliography{references}
\bibliographystyle{IEEEtran}
\newpage

\newpage

\end{document}